\documentclass[runningheads,a4paper]{llncs}

\usepackage{amssymb}
\usepackage{graphicx, booktabs, amsmath}
\usepackage{subfigure}
\usepackage{url}

\begin{document}

\mainmatter  % start of an individual contribution

\title{CASED: Curriculum Adaptive Sampling for Extreme Data Imbalance}

\titlerunning{CASED}

\author{Andrew Jesson \and Nicolas Guizard \and Sina Hamidi Ghalehjegh \and Damien Goblot \and \\ Florian Soudan \and Nicolas Chapados }
% index{Jesson, Andrew}
% index{Guizard, Nicolas}
% index{Hamidi Ghalehjegh, Sina}
% index{Goblot, Damien}
% index{Soudan, Florian}
% index{Chapados, Nicolas}
\authorrunning{Jesson, Guizard, Hamidi, Goblot, Soudan, Chapados}

\institute{Imagia Cybernetics Inc., Montreal, Qc, Canada \\
\email{\{andrew.jesson, nicolas.guizard, sina.hamidi, damien.goblot, \\
florian.soudan, nicolas.chapados\}@imagia.com} \\
\url{http://www.imagia.com}}


\toctitle{Lung Nodule Detection}
\tocauthor{Authors' Lung Nodule Detection}
\maketitle

\begin{abstract}
We introduce CASED, a novel curriculum sampling algorithm that facilitates the optimization of deep learning segmentation or detection models on data sets with extreme class imbalance. We evaluate the CASED learning framework on the task of lung nodule detection in chest CT. In contrast to two-stage solutions, wherein nodule candidates are first proposed by a segmentation model and refined by a second detection stage, CASED improves the training of deep nodule segmentation models (e.g. UNet) to the point where state of the art results are achieved using only a trivial detection stage. CASED improves the optimization of deep segmentation models by allowing them to first learn how to distinguish nodules from their immediate surroundings, while continuously adding a greater proportion of difficult-to-classify global context, until uniformly sampling from the empirical data distribution. Using CASED during training yields a minimalist proposal to the lung nodule detection problem that tops the LUNA16 nodule detection benchmark with an average sensitivity score of 88.35\%. Furthermore, we find that models trained using CASED are robust to nodule annotation quality by showing that comparable results can be achieved when only a point and radius for each ground truth nodule are provided during training. Finally, the CASED learning framework makes no assumptions with regard to imaging modality or segmentation target and should generalize to other medical imaging problems where class imbalance is a persistent problem.
\keywords{lung cancer, computer aided detection, nodule detection, curriculum learning, data imbalance, 3D convolutional neural networks}
\end{abstract}

\section{Introduction}
\label{sec:introduction}

Death rates attributed to lung cancer are three times higher than for any other cancer in the United States \cite{CAAC:CAAC21166}. Diagnosis of this pathology is informed by the presence of malignant pulmonary nodules that appear in thoracic computed tomography (CT) images \cite{Diederich2001}. There is a current trend toward regular monitoring programs of high-risk groups using methods such as low-dose CT \cite{valente2016automatic}. This has been proposed to help catch the pathology in its early stages where, in developed countries, diagnosis dramatically increases the 5-year patient survival rate by 63-75\% \cite{valente2016automatic}. It is likely that radiologists who are tasked with locating and classifying pulmonary nodules would see a dramatic increase in workload with the saturation of such protocols. Fast and accurate automated lung nodule detection methods would then improve lung image evaluation throughput and objectivity by assisting radiologists in their assessment.

One of the major challenges in designing effective automated lung nodule detection methods is the massively unbalanced nature of the data. For example, over the entire Lung Image Database Consortium image collection (LIDC-IDRI) ~\cite{lidc_pub,lidc_data,lidc_tcia} less than 1\% of image voxels contain positive nodule examples. The class imbalance problem has received wide attention in the machine learning and data mining communities, where typical solutions include class over- and under-sampling, weighted losses, and posterior probability recalibration~\cite{5128907}. Sampling schemes have been studied in medical imaging classification (e.g. \cite{pmid24176869} and references therein) and segmentation \cite{Havaei201718}, whereas loss function adjustments were key to results in \cite{Ronneberger2015}. In Computer-Aided Detection (CADe) applications, specialized knowledge can be used, such as limiting the domain of detection to the lung only (requiring a lung masking model) \cite{valente2016automatic}, or training a highly sensitive candidate nodule screening model and then refining predictions by cascading false positive reduction stages \cite{valente2016automatic,MP9562}. A common theme across these approaches is that they tend to be problem-dependent, and sizable efforts must often be expended to find the balancing technique yielding the best performance.

This paper proposes a generic approach to tackle class imbalance, by using, during training, an online adaptation of the distribution of majority and minority class examples, in the spirit of curriculum learning \cite{bengio2009curriculum}. The \emph{Curriculum Adaptive Sampling for Extreme Data imbalance} (CASED) is a novel sampling curriculum that allows for a 3D fully convolutional network (FCN) to yield segmentations high enough in quality to make detection a mere consequence. In contrast to approaches where an off-the-shelf segmentation model~\cite{setio2016pulmonary} or FCN~\cite{long2015fully} is trained to only provide candidates to a second, independently-trained convolutional neural network (CNN) for classification, CASED combines curriculum learning and adaptive data sampling in a way that makes the second classifier redundant. This is achieved by allowing the FCN to first learn how to distinguish nodules from their immediate surroundings while continuously introducing training examples that the model has trouble classifying. This approach yields a surprisingly minimalist proposal to the lung nodule detection problem that tops the LUNA16 challenge~\cite{LUNA16page} leader-board with a score of \emph{88.35\%}. Furthermore, weakly-supervised training, with only a point and radius provided for each training nodule, yields results competitive with those of full segmentation.

\section{Method}
\label{sec:method}
CASED adheres to the observation that the solution to object detection is fully contained in the solution to object segmentation. That is, given an ideal segmentation, a determination of the location, extent, and identity of an imaged object becomes trivial. However, training a model to yield even acceptable medical image segmentations is a considerably harder task than detection for two main reasons. First, manual segmentation of training data is a laborious and expensive endeavour. And second, the model must be able to describe the complex variations of texture ranging over the extent of a given object and its surroundings. Fortunately, the first problem is less significant here as large datasets of annotated lung CT scans are available \cite{lidc_data}; however, robustness to weakly labeled data is important. Regarding the second problem, recent work on FCNs (e.g. FCN-8s for natural images \cite{long2015fully},  U-Net for biomedical images \cite{Ronneberger2015}) has shown that their ability to model multi-scale context over finite image regions makes them ideal candidates for medical image segmentation problems. It behooves one to ask then, in the context of lung nodule detection, why has it not yet been shown that FCNs alone are a competitive solution to this problem? We hypothesize the answer lies in the extreme data imbalance associated to the problem, which has not yet been sufficiently addressed. In the following we present CASED as an approach to overcome this issue.

\paragraph{Curriculum.}
One of the more attractive properties of FCNs is their ability to handle images of arbitrary size. This feature allows us to reduce data imbalance by training on small image patches where the output stride of the model contains at least one positive nodule voxel.  As one would start teaching a child to read the alphabet by restricting their gaze to a large letter A, the model first learns how to represent nodules given only their immediate surroundings. An important consequence of training the FCN on image patches is that we are able to randomize training examples across both patient images and also image regions. Training only on patches that contain nodule examples will result in an extremely sensitive model but with low specificity because it would not learn how to represent the majority of the input image space. Therefore, a curriculum~\cite{bengio2009curriculum} is introduced where the proportion of training patches that contain nodules to those that do not is decreased according to a schedule that tends toward the data distribution as the number of training examples seen approaches infinity. 

\paragraph{Adaptive Sampling.}
After training the FCN using this curriculum with random sampling of background patches, it generally converges to a solution that still gives systematic and predictable false positives. Furthermore, the vast majority of voxels in typical lung images are correctly and confidently predicted as non-nodule, so random sampling would be far more likely to show examples that would have little to no effect on loss optimization. Hence, we introduce a sampling strategy that favours training examples for which prediction using recent model parameters produces false results, an instance of \emph{hard negative mining} (HNM)~\cite{Sung:1998:ELV:275341.275345}. 

\begin{figure}[!t]
\begin{center}
\includegraphics[width=0.5\textwidth]{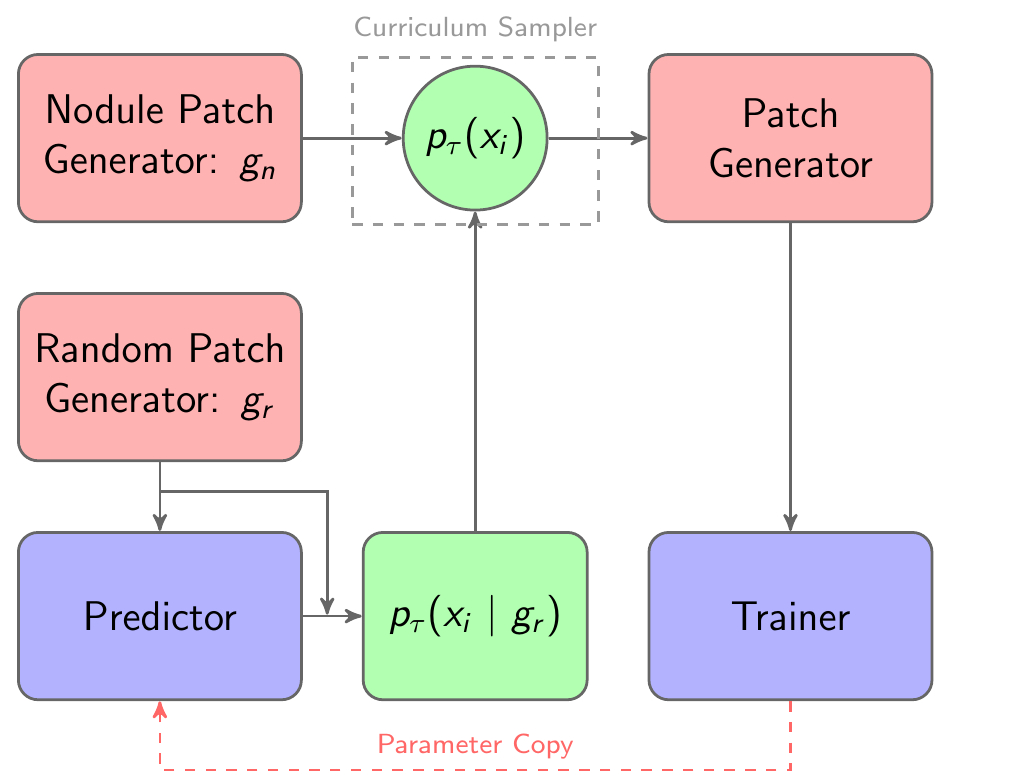}
\end{center}
\caption{Schematic diagram of CASED framework}
\label{fig:schematic_sampler}
\end{figure}

Figure~\ref{fig:schematic_sampler} shows a flowchart of the CASED framework. Let \(\{x_i\}\) be a training set of \(M\) patches. Patch generators are shown in red boxes. The generators \(g_r\) and \(g_n\) represent distributions over the set of all patches and the set of patches that contain nodules, respectively. FCN models are shown in blue boxes where the training model shares its weights with a predictor that is run in parallel for the purposes of HNM. The green boxes represent samplers with distributions that vary with the mini-batch iteration \(\tau\). The sampler \(p_\tau(x_i \mid g_r)\) selects patches based on both \(\tau\) and the training loss \(\mathcal{L}_\tau(x_i)\). The function \(f_r(\mathcal{L}_\tau(x_i), \tau)\) specifying \(p_\tau(x_i \mid g_r)\) must be on the range \([0, 1]\) and \(f_r(\mathcal{L}_\tau(x_i), \tau) \rightarrow M^{-1}\) as \(\tau \rightarrow \infty\). The sampler \(p_\tau(x_i)\) defines the curriculum and chooses between \(g_r\) and \(g_n\) according to a mixing that depends on \(\tau\). The mixing coefficient \(p_\tau(g_n)\) is specified by \(f_n(\tau)\) with range \([0, 1]\) and convergence to \(0\) as \(\tau \rightarrow \infty\). The distribution governing the sampler \(p_\tau(x_i)\) is given by
\begin{equation}
    \begin{split}
        p_\tau(x_i) &= p_\tau(x_i \mid g_r)(1 - p_\tau(g_n)) + p(x_i \mid g_n)p_\tau(g_n) \\
        &= p_\tau(x_i \mid g_r) + \left(p(x_i \mid g_n) - p_\tau(x_i \mid g_r)\right)p_\tau(g_n),
    \end{split}
\end{equation}
\noindent
where \(p(x_i \mid g_n) = 1\) if \(x_i\) contains a nodule, and \(0\) otherwise. In the limit, as \(\tau\) goes to infinity, \(p_\tau(x_i)\) converges to a uniform distribution over \({x_i}\), which makes CASED a valid curriculum \cite{bengio2009curriculum}.

\section{Data and Implementation}
\label{sec:data}
We study CASED as applied to the task of lung nodule detection using the publicly available LIDC image collection \cite{lidc_pub,lidc_data,lidc_tcia}. The LIDC contains 1010 patients and a total of 1018 clinical thoracic CT scans. Each scan has been analyzed through a two-phase nodule annotation process by four expert radiologists. In the first phase each radiologist independently marks nodules as belonging to one of three classes (\emph{nodule $<$ 3mm}, \emph{nodule $\geq$ 3mm}, and \emph{non-nodule $\geq$ 3mm}), where the measurement refers a nodule's diameter. In the second phase, each expert can refine their annotations after seeing the anonymous annotations of the other three radiologists. The LIDC contains 2635 nodules annotated in this way and there are 142 cases that either contain no detected nodules or \emph{nodule \(<\) 3mm}. 

\begin{figure}[t]
\begin{center}
\includegraphics[width=0.8\textwidth]{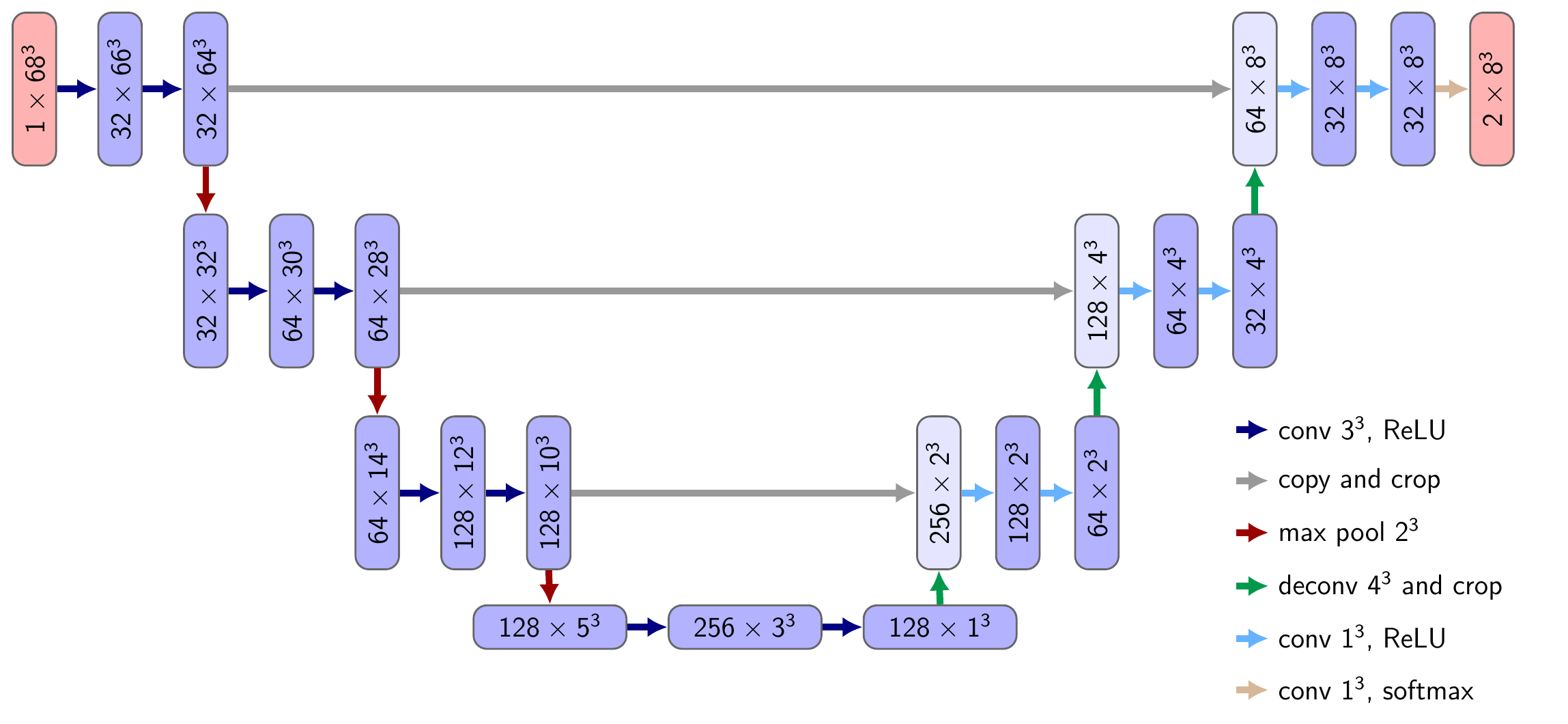}
\end{center}
\caption{Schematic diagram of our 3D U-Net-based architecture.}
\label{fig:unet}
\end{figure}

For segmentation we use a 3D U-Net architecture, based on the model proposed in~\cite{Ronneberger2015}. Figure~\ref{fig:unet} illustrates the model used. The model is comprised of three distinct components: (1)~downstream feature extraction path, (2)~upstream feature pooling path, and (3)~linear pixel classifier. In the downstream path, we use layers of ``convolution'' and ``pooling''. Each layer effectively encodes a progressively larger image neighbourhood of the input image as we go deeper. In the upstream path, we use layers of ``convolution'' and ``strided transposed convolution'' layers. Multi-scale features extracted in the downstream path are combined to provide pixel-level features in the input image space. Finally, the linear pixel classifier uses a simple ``sigmoid'' layer to provide per-pixel prediction of nodule or non-nodule.

CASED training requires minimal data preprocessing. For a given CT scan, image intensities are transformed to Hounsfield units and linearly rescaled.
%such that $-1250$\emph{HU} maps to $-1$ and $250$\emph{HU} maps to $1$. 
The scan is then resized to 1.25\emph{mm} isotropic voxels. For training, binary segmentation maps are built from the expert annotations listed in the provided XML files and are also transformed into the 1.25\emph{mm} isotropic space. The binary segmentation maps are nodule-wise refined to only label as nodule those voxels that correspond to the intersection of all available annotations. For example, if a nodule only has an annotation from one rater, that annotation is used; however, if a nodule has annotations from multiple raters the intersection of those annotations is used.

Training is done by optimizing voxel-wise binary cross-entropy over each prediction patch (of size $8^3$) and its corresponding reference segmentation using stochastic gradient descent with Nesterov momentum. We use mini-batches with 16 image patches of size $68^3$ as input. Nodule patches are defined as those for which there is a labeled nodule voxel within the $8^3$ output stride. All other patches are called background. The curriculum is initialized with \(p_\tau(g_n) = 1.0\) and is decayed after each mini-batch iteration. Finally, ``background'' patches are sampled based on whether they contain a false positive prediction using recent model parameters. 

At test time, an equally minimalist approach to postprocessing is required. Given a test image, the model outputs a soft segmentation map estimating the probability that a given voxel belongs to the nodule class. This map is thresholded giving a binary segmentation on which connected component analysis is performed to yield candidate nodules. The center of mass and average value of the segmentation map over each candidate is found to yield a list of point and confidence predictions. The points are finally transformed back into the native image space. Because the model is fully convolutional, the input size at test need only be divisible by 8. Given sufficient GPU memory the entire CT scan can be passed as input without tiling and full prediction takes only a few seconds.

\section{Experiments and Results}
\label{sec:experiments}

We evaluate the CASED framework using the 2016 Lung Nodule Analysis Challenge (LUNA16) 10-fold cross-validation split \cite{LUNA16page}. Each fold contains 88-89 CT scans. The reference standard for LUNA16 consists of all \emph{nodule $\geq$ 3mm} that have been detected by at least three of four raters. Evaluation is based on the detection sensitivity at various false positive rates per scan. A detailed explanation of the evaluation can be found on the LUNA16 website \cite{LUNA16page}.

\begin{table*}[t]
\label{tb:luna_candidate}
\renewcommand{\arraystretch}{1.3}
\setlength{\tabcolsep}{6pt}
\caption{The LUNA16 cross-validation sensitivity at different number of false positives per scan. The scores for other methods are taken from the result section of LUNA16 website \cite{LUNA16page}. Method with asterisk superscript does not provide any description on LUNA16 scoreboard.}

\begin{center}
\scriptsize
\begin{tabular}{lcccccccc}
\toprule
& \multicolumn{7}{c}{False Positives Per Scan} & \\
Method                                      & 0.125 & 0.25 & 0.5 & 1 & 2 & 4 & 8	& \textbf{Average} \\
\midrule
\midrule
CASED                                      & 0.759 & 0.825 & 0.866 & 0.903 & 0.926 & 0.946 & 0.959 & \textbf{0.883} \\
%CASED-1                                       & 0.727 & 0.805 & 0.847 & 0.895 & 0.915 & 0.938 & 0.954 & 0.869\\
%CASED-2                              & 0.739 & 0.811 & 0.861 & 0.891 & 0.919 & 0.940 & 0.950 & 0.873 \\
CASED-Sphere                                & 0.701 & 0.762 & 0.813 & 0.862 & 0.897 & 0.923 & 0.939 & 0.842 \\
\midrule
AIDENCE\textsuperscript{*}                                     & 0.739 & 0.788 & 0.828 & 0.879 & 0.910 & 0.938 & 0.963 & 0.864 \\
%Ethan20161221\textsuperscript{*}                               & 0.667 & 0.780 & 0.842 & 0.887 & 0.925 & 0.940 & 0.950 & 0.856 \\
ZNET \cite{setio2016validation}             & 0.661 & 0.724 & 0.779 & 0.831 & 0.872 & 0.892 & 0.915 & 0.811 \\
ETROCAD \cite{etrocad,setio2016validation} & 0.250 & 0.522 & 0.651 & 0.752 & 0.811 & 0.856 & 0.887 & 0.676 \\
M5LCAD \cite{m5l,setio2016validation}  & 0.306 & 0.360 & 0.540 & 0.691 & 0.762 & 0.797 & 0.798 & 0.608 \\
\bottomrule
\end{tabular}
\label{tb:luna_sensitivity}
\end{center}
\end{table*}

For each test fold, we train on eight and validate on one of the remaining folds. We also use model ensembling to improve the reliability of the results. Finally, we repeat the experiment using spherical segmentations defined by the location and radius of each nodule instead of the reference annotations (CASED-Sphere). 

\begin{figure}[!t]
\centering
\begin{subfigure}
  \centering
  \includegraphics[height=.26\textheight]{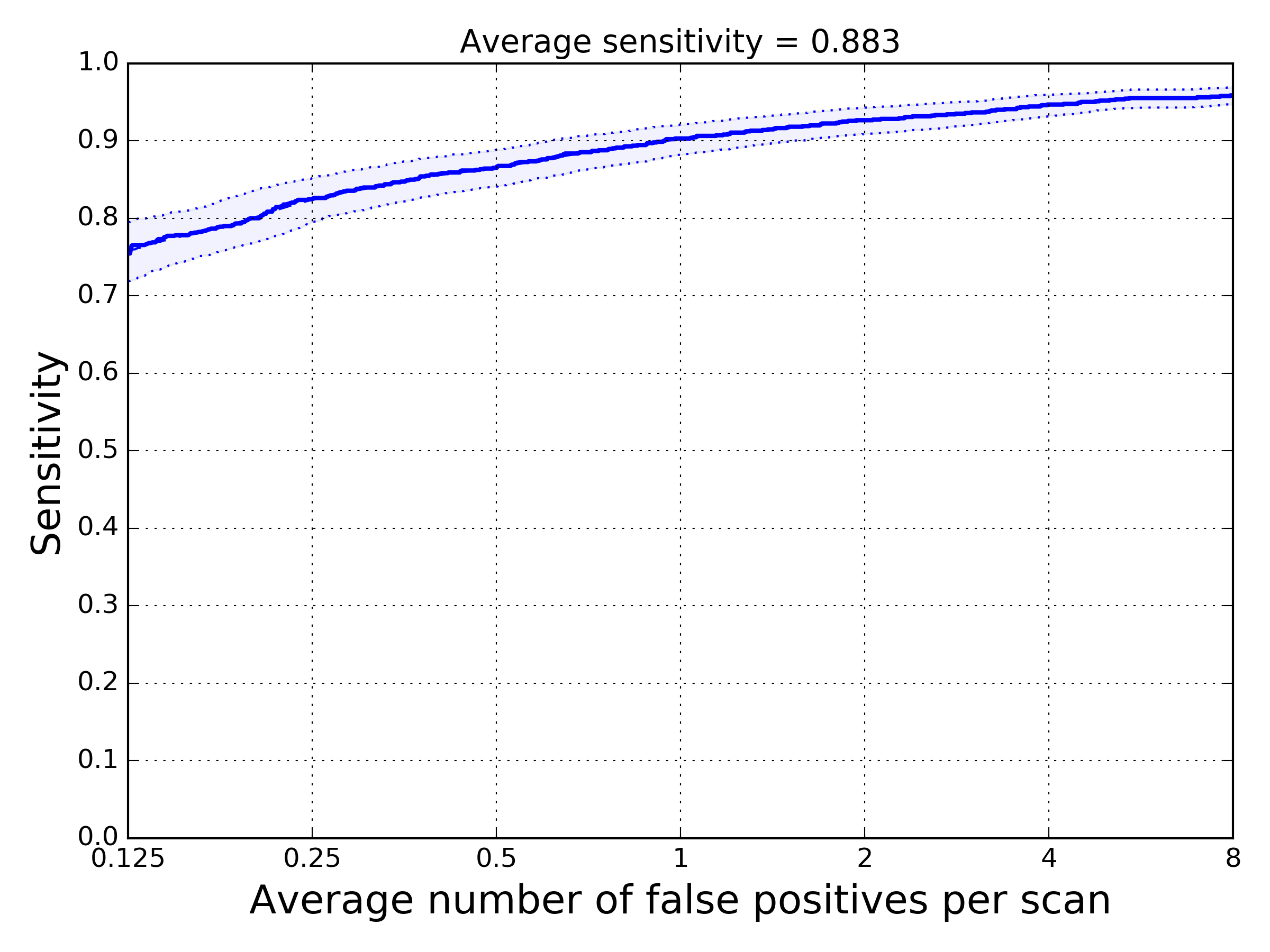}
  \label{fig:lunasub1}
\end{subfigure}%
\hspace*{5mm}%
\raisebox{0.5mm}{%
\begin{subfigure}
  \centering
  \includegraphics[height=.24\textheight]{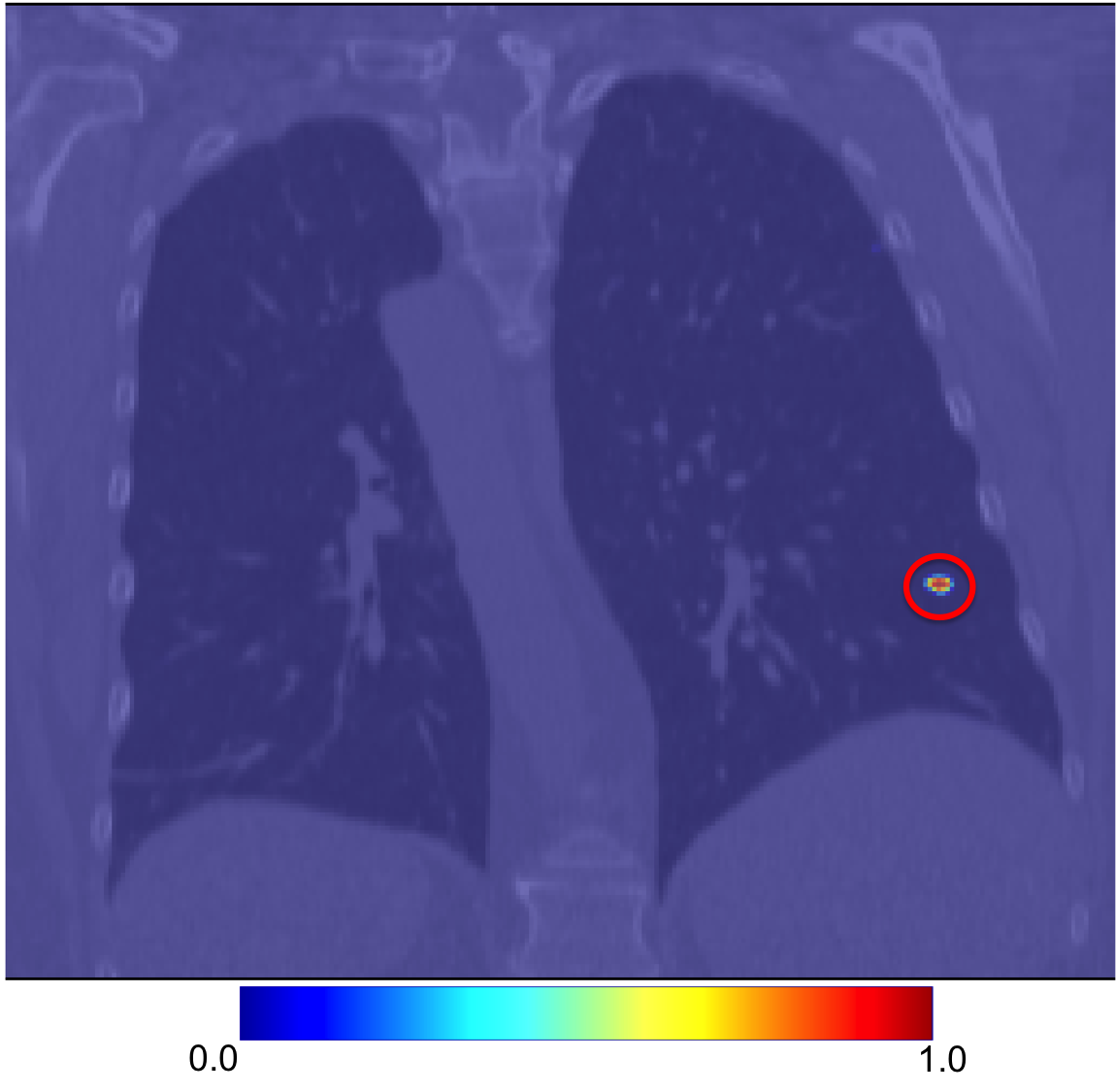}
  \label{fig:exemplesub2}
\end{subfigure}}
\caption{\textbf{Left:} The free-response receiver operating characteristic (FROC) curve for CASED. The blue line and shaded area represent the mean and variance of the nodule detection sensitivity over 1000 bootstrapped samples at different false positive rates. \textbf{Right:} Lung CT overlaid by probability map. In the color spectrum, as we move toward right (red) the probability of being nodule increases.}
\label{fig:luna}
\end{figure}

Table~\ref{tb:luna_sensitivity} summarizes the results of these experiments for the lung nodule detection task and provides a comparison to the results of other methods submitted to the LUNA16 leader board. The CASED learning framework shows a 8.9\% relative increase in average sensitivity over the best published results for a given model, ZNET \cite{setio2016validation}. The free-response receiver operating characteristic (FROC) curve for CASED appears in Figure~\ref{fig:luna}. Finally, we demonstrate robustness to segmentation quality by showing that a 3.8\% relative increase over ZNET is achieved with CASED-Sphere. 

\section{Conclusions}
\label{sec:conclusion}
This paper proposes CASED, a new curriculum sampling algorithm for the highly class imbalanced problems that are endemic in medical imaging applications. We demonstrate that CASED is a robust learning framework for training deep lung nodule detection models. Evaluated on the LUNA16 challenge, we achieve the current state-of-the-art leader-board performance with an average sensitivity score of 88.35\%. Since the CASED algorithm does not require any assumption on image modality, it can be applied to any arbitrarily large dataset wherein the unbalanced nature of data poses major problems for designing automated methods.

\bibliography{ref}
\bibliographystyle{splncs03}
\end{document}